\documentclass[runningheads]{llncs}
\usepackage{graphicx} 
\usepackage{hyperref}
\usepackage{xcolor}
\usepackage{subcaption}
\usepackage{amsmath}
\usepackage{amsfonts} 
\usepackage{orcidlink}
\usepackage{siunitx} 

\newcommand{\VBS}{\text{VBS}_\text{full}}
\newcommand{\SBS}{\text{SBS}_\text{full}}
\newcommand{\PIAS}{\text{PIAS}_\text{perf}}
\newcommand{\ELAperf}{\text{ELA}_\text{perf}}

\newcommand{\ELAbudget}{B_\text{ELA}}
\newcommand{\VBSopt}{\text{VBS}_\text{opt}}

\makeatletter
\newcommand{\printfnsymbol}[1]{%
  \textsuperscript{\@fnsymbol{#1}}%
}
\makeatother

\title{On the Influence of the Feature Computation Budget on Per-Instance Algorithm Selection for Black-Box Optimization}
\titlerunning{On the Influence of the Feature Computation Budget on PIAS for BBO}

\author{Koen van der Blom\thanks{Both authors contributed equally.}\inst{1}\orcidlink{0000-0002-4653-0707} and Diederick Vermetten\printfnsymbol{1}\inst{2}\orcidlink{0000-0003-3040-7162}}
\authorrunning{Koen van der Blom and Diederick Vermetten}
\institute{
Centrum Wiskunde \& Informatica, Amsterdam, The Netherlands\\
\email{kvdb@cwi.nl}
\and
Sorbonne Universit\'e, CNRS, LIP6, Paris, France\\
\email{diederick.vermetten@lip6.fr}
}

\begin{document}

\maketitle

\begin{abstract}
Per-instance algorithm selection (PIAS) takes advantage of complementarity between a set of algorithms by deciding which algorithm to run on a given instance. This decision is based on features of the instances, which, in the context of black-box optimization (BBO), require a part of the optimization budget to be computed.
This raises two questions: (a) from which fraction of the budget spent on feature computation does PIAS become worth it for BBO, and (b) which fraction of the budget optimizes the tradeoff between feature accuracy and PIAS performance. To this end, we perform a broad study where PIAS with varying sampling budgets for feature computation is compared to the single best algorithm on a broad range of algorithm selection scenarios. These scenarios consist of two portfolio sizes, three problem sets, 4 dimensionalities, and 10 target budgets.
We find that PIAS is viable for the majority of tested scenarios, even when as much as a quarter of the total budget is spent on feature computation. The tradeoff for the fraction of the budget spent on feature computation to maximize the benefit of PIAS is highly dependent on the specific AS scenario.
Further, on average 20 percent of PIAS loss to the virtual best solver is explained by the budget spent on feature computation, highlighting the importance of properly accounting for the feature budget.
\end{abstract}

\section{Introduction}

For most problems a variety of algorithms is available to solve them, but it is not obvious which one is the best. When there are many instances of the problem, and there is complementarity between the algorithms in terms of for which instances they perform best, this becomes a recurring question.
This has resulted in substantial attention for the algorithm selection (AS) problem~\cite{Rice1976} across different domains, such as SAT~\cite{XuEtAl2008}, MIP~\cite{XuEtAl2011}, and black-box optimization~\cite{BischlEtAl2012} (BBO).

At its core, AS is a learning problem, where given a new problem instance, we aim to select a single method from a range of possibilities. A key question thus arises in the representation of the problem. For per-instance algorithm selection (PIAS), individual problem instances are typically represented by a set of instance features. In some contexts, these features are inherent from the problem formulation, or can be extracted without incurring a high cost. For example, in SAT solving, there are features that can be computed directly from the instance file~\cite{xu2012features}. However, when dealing with black-box optimization (BBO), the only a priori available information about an instance is the shape of the search space, e.g., the number and types of decision variables. This is insufficient to differentiate between individual problem instances.

The introduction of exploratory landscape analysis~\cite{MersmannEtAl2011} (ELA) provided features that aim to describe the optimization landscape for BBO problems by sampling the objective function.

In turn, this allowed PIAS to be applied in the context of BBO by using part of the total evaluation budget to sample solutions to compute ELA features, while using the remaining budget for the selected algorithm~\cite{BischlEtAl2012,KerschkeTrautmann2019}.

A common baseline for PIAS is the so-called single best solver (SBS), representing the algorithm that performs the best on aggregate across all relevant instances. In many settings, the SBS and the algorithm chosen by PIAS are given the same budget, because the cost of feature computation is negligible. For BBO, however, this cost may be substantial, and, for a fair comparison, the budget for the selected algorithm should be reduced by the budget used for feature computation. The performance of the solutions sampled for feature computation should also be counted towards the selector performance. Although the effect of this may quickly diminish for larger total budgets, it may substantially affect performance in low-budget scenarios. Unfortunately, not all papers in this area make clear whether they account for these factors or not.

In this paper, we compare PIAS and the SBS for various feature computation budgets, in order to investigate the tradeoff between the improved reliability of features when a larger fraction of the total budget is used to compute them, and the loss in performance due to the selected optimizer being left with a smaller budget. By studying a wide variety of AS scenarios with different problem sets, algorithm portfolios, total budgets, and dimensionalities, we aim to provide a more general perspective. Specifically, we aim to answer the following two questions:

\textbf{Research question A:} When is it worth using a fraction of the optimization budget to apply per-instance algorithm selection in black-box optimization?

\textbf{Research question B:} What is the optimal fraction of the total budget to spend on feature computation?

We show that PIAS can reliably be used for most scenarios, even when as much as \SI{25}{\percent} of the budget is spent on feature computation. However, the optimal tradeoff varies greatly based on the exact settings of the considered algorithm selection scenario. Finally, we observe that for PIAS on average \SI{11}{\percent} to \SI{28}{\percent} (depending on the problem set) of the total loss comes from budget spent on feature computation, highlighting the importance of properly accounting for the feature budget.

\section{Background and Definitions}

To provide the appropriate background, we briefly introduce key concepts in algorithm selection, and notation for the budgets and methods we compare.

The algorithm selection problem~\cite{Rice1976} considers that for a set of problem instances $\mathcal{I}$, there is often no single algorithm $A$ in a portfolio $\mathcal{A}$ that performs the best on every problem instance. This problem is relevant across domains, but our focus is on BBO. There are some review papers that discuss the particular challenges of PIAS for BBO, but they were done when few practical implementations existed~\cite{KerschkeEtAl2019,MunozEtAl2015b}. Since then, algorithm selection for BBO has received increased attention primarily in the single-objective continuous setting, including the use of deep-learning models~\cite{AlissaEtAl2023,SeilerEtAl2025} and dynamic algorithm selection~\cite{KostovskaEtAl2022}, but has also branched out to, e.g., constrained multi-objective optimization~\cite{AndovaEtAl2026}. In terms of features that can be used, a recent survey covered the available feature sets for continuous single-objective BBO~\cite{cenikj2026survey}.

The algorithm selection problem can be formulated as the question of how we can select the best algorithm $A\in\mathcal{A}$ for each instance $i\in\mathcal{I}$, such that we optimize a performance metric $m: \mathcal{A} \times \mathcal{I} \rightarrow \mathbb{R}$. To assess the quality of an algorithm selector, two baselines are used most commonly: the single best solver (SBS) and the virtual best solver (VBS). The SBS represents the algorithm that performs the best on aggregate across all instances in $\mathcal{I}$, whereas the VBS represents the hypothetical perfect selector that, for each individual instance in $\mathcal{I}$, chooses the best algorithm in $\mathcal{A}$.

\begin{figure}
    \centering
    \includegraphics{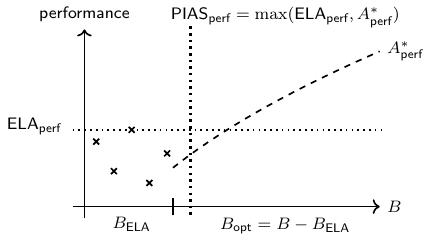}
%
%
%
    \caption{Schematic visualization showing how the budget is split into feature computation and optimization, and how this relates to selector performance.
    }
    \label{fig:budget_split}
\end{figure}

Since for BBO, spending part of the evaluation budget on feature computation is a requirement for PIAS, we have to account for this when evaluating PIAS methods. To this end, we define the full evaluation budget $B$, the budget spent on sampling for ELA $\ELAbudget$, and the remaining budget available for optimization after sampling $B_\text{opt}=B-\ELAbudget$ (see Figure~\ref{fig:budget_split} for a visual representation of this).

These different budgets in turn influence the performance evaluation of PIAS methods. We define: 
\begin{itemize}
    \item $\ELAperf$ as the best performance found based on the samples used for ELA with budget $\ELAbudget$;
    \item $A^*_\text{perf}$ as the best performance found by the algorithm $A^*\in\mathcal{A}$ chosen by PIAS with budget $B_\text{opt}$;
    \item and $\PIAS$ $=\max(\ELAperf,A^*_\text{perf})$ as the performance achieved by PIAS, accounting for both the points sampled for ELA, and the performance of the chosen algorithm, i.e., considering the full budget $B$.
\end{itemize}

Combined with this, we consider the following standard baselines, adjusted to explicitly take into account the BBO-specific complexities:
\begin{itemize}
    \item Single best solver ($\SBS$): The algorithm with the best mean performance over all \emph{training} instances when using the \emph{full budget} $B$. This represents the best choice that could be made without using PIAS.
    \item Virtual best solver for the full budget ($\VBS$): The best algorithm per instance when using the full budget $B$.
    \item Virtual best solver for the optimization budget ($\VBSopt$): The best algorithm per instance when using the remaining budget after feature computation $B_\text{opt}$.
\end{itemize}

Finally, we note that the performance of an algorithm selector can be measured by looking at the fraction of the so-called VBS-SBS gap that it closes. For this measure, we use the full-budget versions for both VBS and SBS, as this most closely matches the `oracle' perspective of the VBS. However, in some cases, $\PIAS$ might outperform $\VBS$ if the sampling used for feature calculation finds a better solution than the best algorithm in the portfolio, resulting in a gap-closed measure $>1$.

\section{Experimental Setup}

Since the performance of algorithm selection in continuous black-box optimization settings is highly dependent on both the used problems and algorithms, we make use of a variety of problem suites and algorithm sets in our experiments. From the problem side, we select three sets of problem instances:
\begin{itemize}
    \item \textbf{BBOB}~\cite{HansenEtAl2009}: As the most commonly used benchmark suite for unconstrained single-objective, noiseless optimization, BBOB has naturally been the default for algorithm selection studies~\cite{BischlEtAl2012}. While its choice as a selection testbed has been criticized recently~\cite{kononova2025benchmarking},
    we opt to include it as a setting where PIAS is likely to perform well, and where deeper analysis of the results is possible, by considering, e.g., for which specific problems the selector succeeds or fails.
    \item The \textbf{MA-BBOB} generator~\cite{VermettenEtAl2025}: Based on affine recombinations of the base BBOB problems~\cite{DietrichMersmann2022}, this represents a scenario without the built-in clusters of instances found in BBOB.
    \item The \textbf{RandOptGen} (ROG) generator~\cite{SeilerEtAl2025b}: This generator represents an alternative scenario without built-in clusters of instances, that is not based on BBOB, but rather creates problems using a tree-based approach.
\end{itemize}

For each of these problem sets, we consider problem dimensionalities $d\in\{2,5,10,20\}$, and for each dimensionality, we generate a number of instances. For BBOB, we use $10$ instances for each of the $24$ problems, while for the generators, we generate $250$ instances at random. 

For our algorithm portfolio, we make use of a set of 22 optimizers from the Nevergrad platform~\cite{RapinTeytaud2018}.\footnote{Algorithms considered:
RCobyla, DiagonalCMA, NelderMead, OnePlusOne, OnePlusLambda, RLSOnePlusOne, RealSpacePSO, NaiveIsoEMNA, RFMetaModelPSO, DE, RSLSQP, Powell, ManyLN, QODE, CMAsmall, CMAbounded, MultiCMA, CMA, Cobyla, MetaModel, MetaModelOnePlusOne, PMetaModel(CmaFmin2).} Note that although there is diversity and complementarity among the considered algorithms, the selection is also somewhat arbitrary. This is representative of a realistic scenario, where (exact) relations and complementarity between available algorithms is unknown.
We run each algorithm for a total of $2500d$ evaluation, but we construct different algorithm selection settings by taking shorter prefixes of these runs. The different budget settings we consider then become: $B\in\{10d, 15d, 25d, 50d, 100d, 250d, 500d, 1000d, 1500d, 2500d\}$. 

Finally, to zoom in on the relative benefits of using some fraction of the budget for feature computation, we use the following feature computation budgets:\\ $\ELAbudget\in\{5d, 10d, 25d, 50d, 100d, 250d\}$, where we ensure $\ELAbudget<B$.

For every algorithm, we use $5$ independent evaluations for each problem instance. We make use of IOHexperimeter~\cite{deNobelEtAl2024} for logging and IOHinspector~\cite{vermetten2025mo} for processing the performance data. We use the box-constrained versions of all problems, with bounds $[-5,5]^d$ for BBOB and MA-BBOB and $[-1,1]^d$ for ROG, following their respective default settings. The full performance data, as well as full reproducibility instructions and additional figures which could not be included here are made available on our Zenodo repository~\cite{zenodo}. 

\subsection{Portfolio Size}

Since the size of the considered algorithm portfolio can play an important role in the performance of algorithm selection~\cite{kostovska2023ps}, we split our settings into two categories. First, we consider the full portfolio of $22$ optimizers. Second, we consider portfolios of size $4$, selected to have high \textbf{algorithm complementarity}. We use algorithm complementarity as a measure of the potential benefit of algorithm selection, and measure it as the difference between the VBS and SBS performance. Note that this is not intended to create optimal algorithm selection performance (as in~\cite{kostovska2023ps}), but scenarios in which we can better demonstrate the impacts of the different components of the algorithm selection pipeline.

With this definition of algorithm complementarity, we could find a subset to maximize it for each setting by searching through all possible subsets. However, this is computationally intensive, and since high but non-optimal complementarity is sufficient here,
we opt for a heuristic procedure instead. This is a two-stage approach, where we first estimate Shapley values~\cite{Shapley1953} for each algorithm. To this end, we use $\VBS-\SBS$ as our target measure, and use the marginal contribution of each algorithm to a random set of portfolios to estimate how much each algorithm helps to improve the overall complementarity. These approximated Shapley values are then normalized and used to guide a search for the final portfolio. We use a random search with weighted probabilities for each algorithm based on their Shapley value to find the portfolio of size $4$ that maximizes $\VBS-\SBS$. Separate size-4 portfolios are created for each combination of problem set, budget factor, and dimensionality.

\subsection{Algorithm Selection Procedure}

\paragraph{Feature Computation} 
While there are a large variety of feature sets available for black-box optimization problems~\cite{cenikj2026survey}, we opt to use ELA, which is most commonly used. Given a budget for feature computation, we use Sobol sampling to get function values, normalize them~\cite{prager2023nullifying}, and calculate the ELA features using the \texttt{pFlacco} package~\cite{prager2024pflacco}. We then perform a filtering stage, where we remove any features that resulted in NaN or infinite values 
for any instance or are flat across all instances. For each problem instance, feature computation is repeated five times to account for the variance in feature values that may result from different samples.

\paragraph{Performance Measure}
For our performance measure, we use normalized values in $[0,1]$ to simplify comparisons. For BBOB and MA-BBOB, we make use of the empirical attainment function~\cite{lopez2024using}, with bounds $10^2$ and $10^{-8}$ with a logarithmic scaling between them, following conventional practices from COCO~\cite{hansen2021coco}. For ROG, we do not have known optima, so we instead perform a standard normalization per instance using the extrema found by all algorithms between $5d$ and $2500d$ evaluations.\footnote{One ROG instance for $d=5$ is excluded from the selector experiments, because function values were too small to reliably measure performance differences.}

\paragraph{Cross-Validation} 
Throughout our experiments, we employ 5-fold cross-validation, where we train on 4 folds and test on the remaining fold. For the split between folds, we use the instance ID, following the so-called leave-instance-out~\cite{BischlEtAl2012} strategy. We do not consider another popular strategy called leave-problem-out~\cite{BischlEtAl2012}, which splits by problem. The reason for this is that the leave-problem-out strategy tests generalization beyond the training distribution~\cite{CenikjEtAl2025b}, which is not the topic of this paper.

\paragraph{Selection Model}
To construct an algorithm selector, we train a multi-output \texttt{RandomForestRegressor} from the \texttt{scikit-learn}~\cite{PedregosaEtAl2009} package.
This method was selected for its simplicity, and because earlier work suggests that the type of machine learning model used matters little~\cite{kostovska2023comparing}. Although it is not clear how those results generalize beyond BBOB, for the purposes of this paper a simple selector is sufficient.
For the instances in the testing set, the algorithm with the highest predicted performance is selected. For both the training and the testing sets, the mean over all 5 repetitions is used, and matched with each repetition of the feature computation.

\section{Results}

By varying the problem sets, problem dimensionality, algorithm portfolios, and budgets for optimization and feature calculation, we create a wide range of in total \num{1440} different algorithm selection scenarios. We note that, in principle, each combination represents an independent AS scenario, and care must be taken to avoid drawing too general conclusions based on apparent trends or correlations between results on different scenarios.

\begin{figure}[t]
    \centering
    \includegraphics[width=0.95\linewidth]{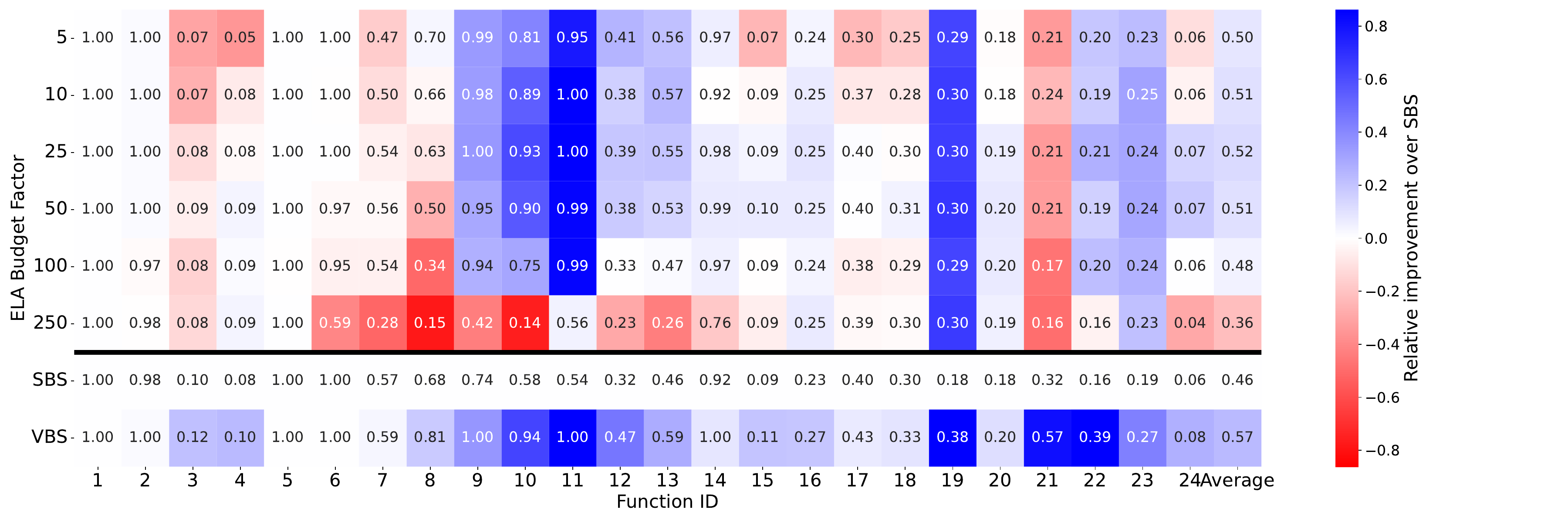}
    \caption{PIAS performance when trained with different ELA budget factors, compared to the $\SBS$ and $\VBS$ baselines. Each column shows one of the 24 BBOB problems, with results averaged across instances and folds. The total budget factor is $500$, $d=10$, and the portfolio size is 22. Color indicates relative difference to the $\SBS$, while the numbers indicate the raw performance in terms of attainment.}
    \label{fig:BBOB_single_scenario}
\end{figure}

\subsection{BBOB Results}

We start our analysis of the impact of the feature computation budget by looking at the BBOB problem suite in combination with the full algorithm portfolio. By looking at a single algorithm selection scenario, selector performance can be visualized on a per-function basis (averaged over instances), as shown in Figure~\ref{fig:BBOB_single_scenario}. When also averaging over the functions (shown in the last column in this figure), the tradeoff in terms of feature budget becomes apparent. Performance initially increases as more evaluations are allocated to the ELA part of the selection procedure, but when this factor becomes too large relative to the total budget, it is detrimental to overall performance. This is also visible on individual functions, for example, F10, where the highest ELA budget leads to a substantial drop in performance of the selected algorithm. In this case, using a feature budget of $250d$ leads to a PIAS performance which is clearly below that of the SBS. 

\begin{figure}[h]
    \centering
    \includegraphics[width=0.95\linewidth]{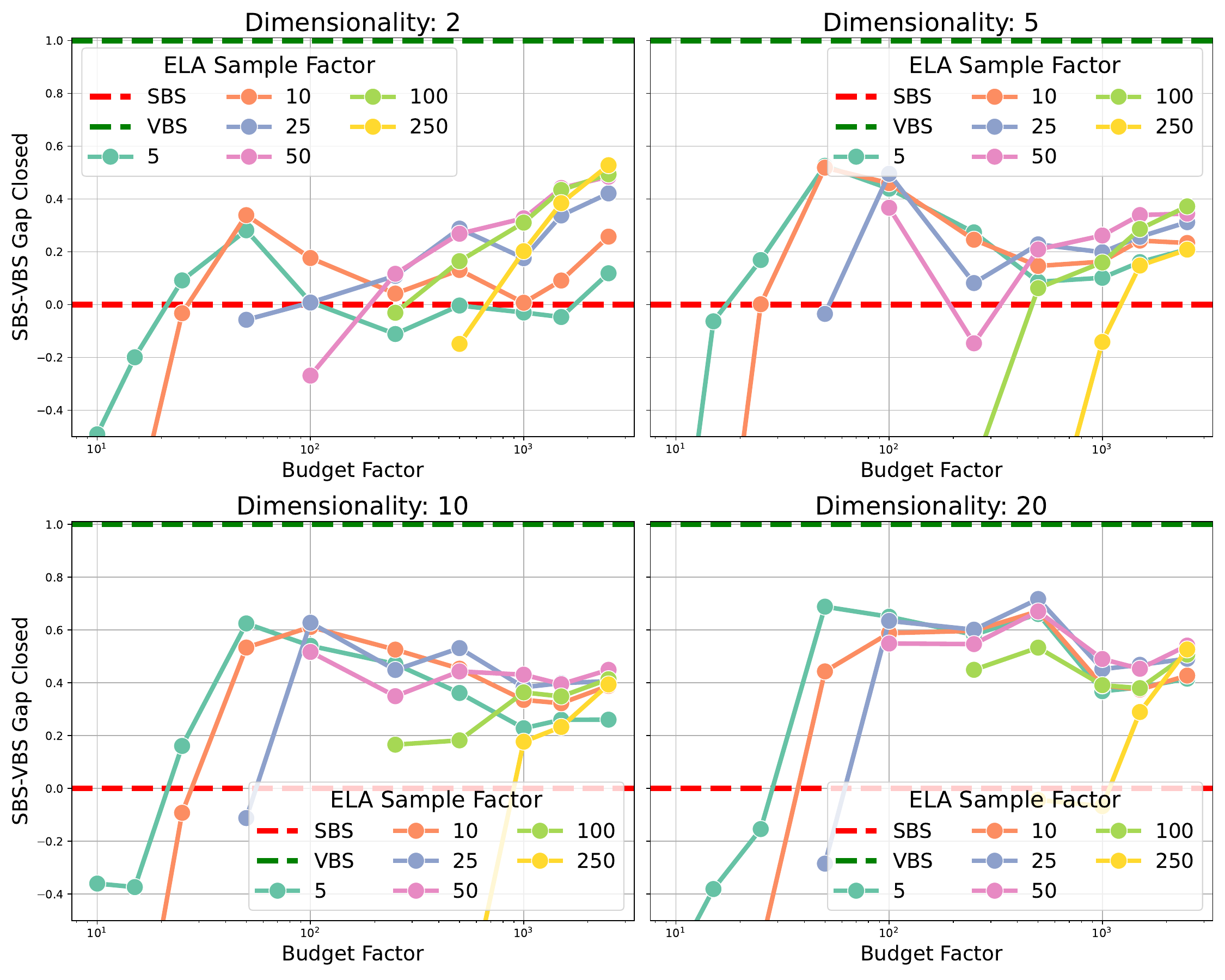}
    \caption{Fraction of the SBS-VBS gap closed (mean over 5 folds) by PIAS for different total budgets, separated by how much budget is spent on feature computation. Each subplot represents a different dimensionality of the BBOB problems, and all use the full algorithm portfolio.
    }
    \label{fig:BBOB_gapclosed}
\end{figure}

Where Figure~\ref{fig:BBOB_single_scenario} provides a detailed view of a single algorithm selection scenario, we can aggregate its contents in order to analyze the impact of relative feature budget for a wider range of total budget values. We can subsequently use the fraction of the VBS-SBS gap closed by the selector as the performance measure. With this performance measure, positive values indicate that PIAS improves over the SBS. In Figure~\ref{fig:BBOB_gapclosed}, we show the gap closed by PIAS for the BBOB scenarios with the full algorithm portfolio, for each of the 4 selected problem dimensionalities. 

From Figure~\ref{fig:BBOB_gapclosed}, we can see that the trend observed in Figure~\ref{fig:BBOB_single_scenario} seems to hold across most settings: when the feature calculation budget becomes too large relative to the total budget, the selection tends to perform rather poorly. Similarly, we note that the lower feature budgets are insufficient for the smallest total budgets, then become more beneficial, and drop off again relative to larger feature budgets as the total budget increases even further. This shows the inherent tension in allocating a fraction of the budget to feature computation, where too large a fraction leads to poor performance, but too small a fraction is suboptimal for larger total budgets. Although we do not show the results for the portfolio with 4 algorithms on the BBOB suite, in almost all cases, PIAS worked even better.

\subsection{Other Problem Suites}
\begin{figure}[h]
    \centering
    \includegraphics[width=0.95\linewidth]{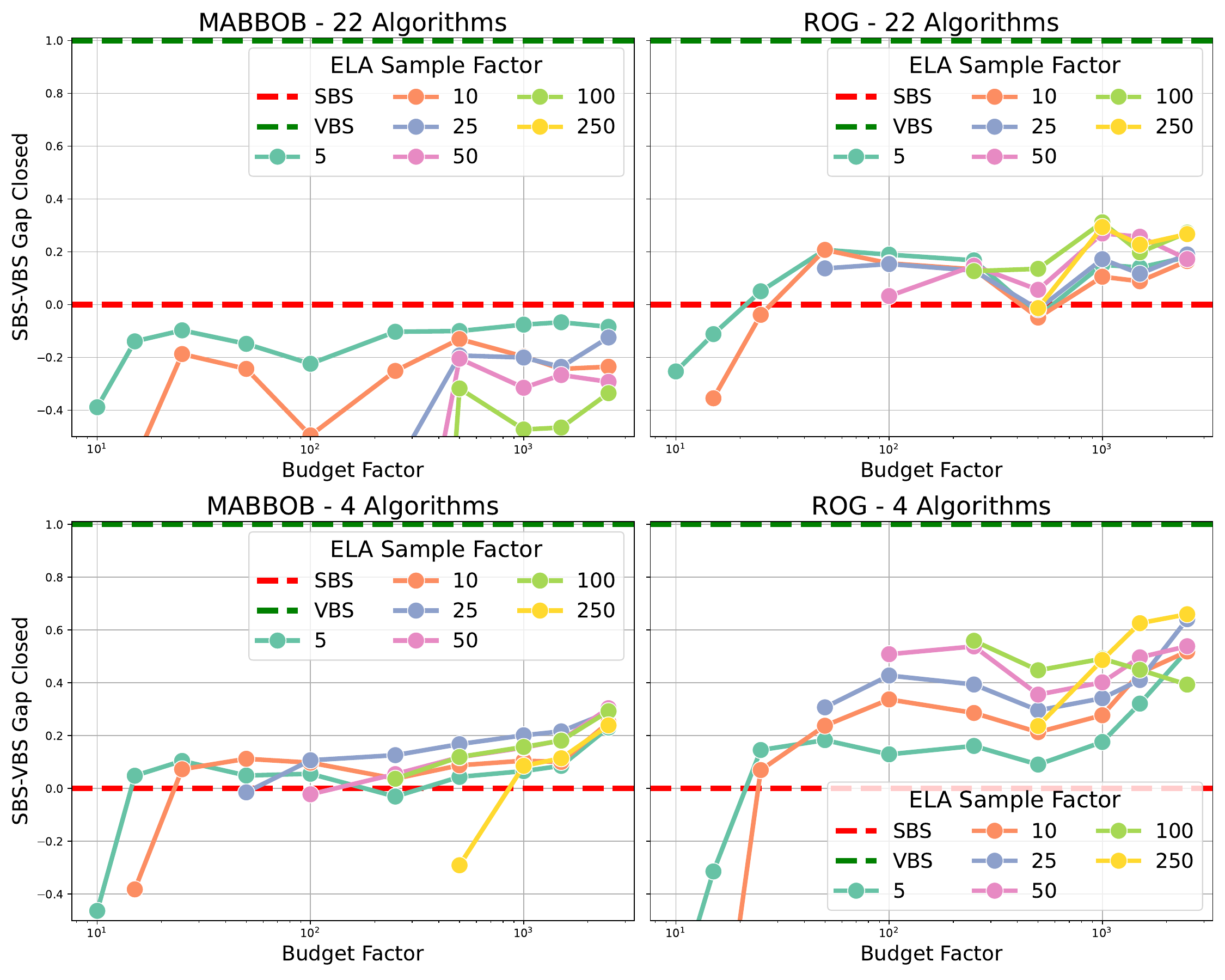}
    \caption{Fraction of the SBS-VBS gap closed (mean over 5 folds) by PIAS for different total budgets, separated by how much budget is spent on feature computation. Each subplot represents a different problem and algorithm set, as indicated in its title. All plots are using problem dimensionality 5.}
    \label{fig:Other_gapclosed}
\end{figure}

We observe that in general, per-instance algorithm selection shows clear benefits over the SBS on the BBOB set when some minimal requirements are met in terms of sampling and total budget. This matches previous observations on BBOB when evaluating PIAS with the leave-instance-out approach~\cite{renau2021towards}, which is why it is generally considered a rather simple algorithm selection scenario. To validate whether our findings translate to other problem suites, we show the same gap-closed measure for the MA-BBOB and ROG problems with $d=5$ in Figure~\ref{fig:Other_gapclosed}. In the first row of this figure, we see the scenarios with the full algorithm portfolio. Here, it is notable that the MA-BBOB scenarios consistently lose out relative to the SBS baseline. In the second row, where the algorithm portfolio has a reduced size, the benefits of PIAS become evident. 

\begin{figure}[t]
    \centering
    \includegraphics[width=0.95\linewidth]{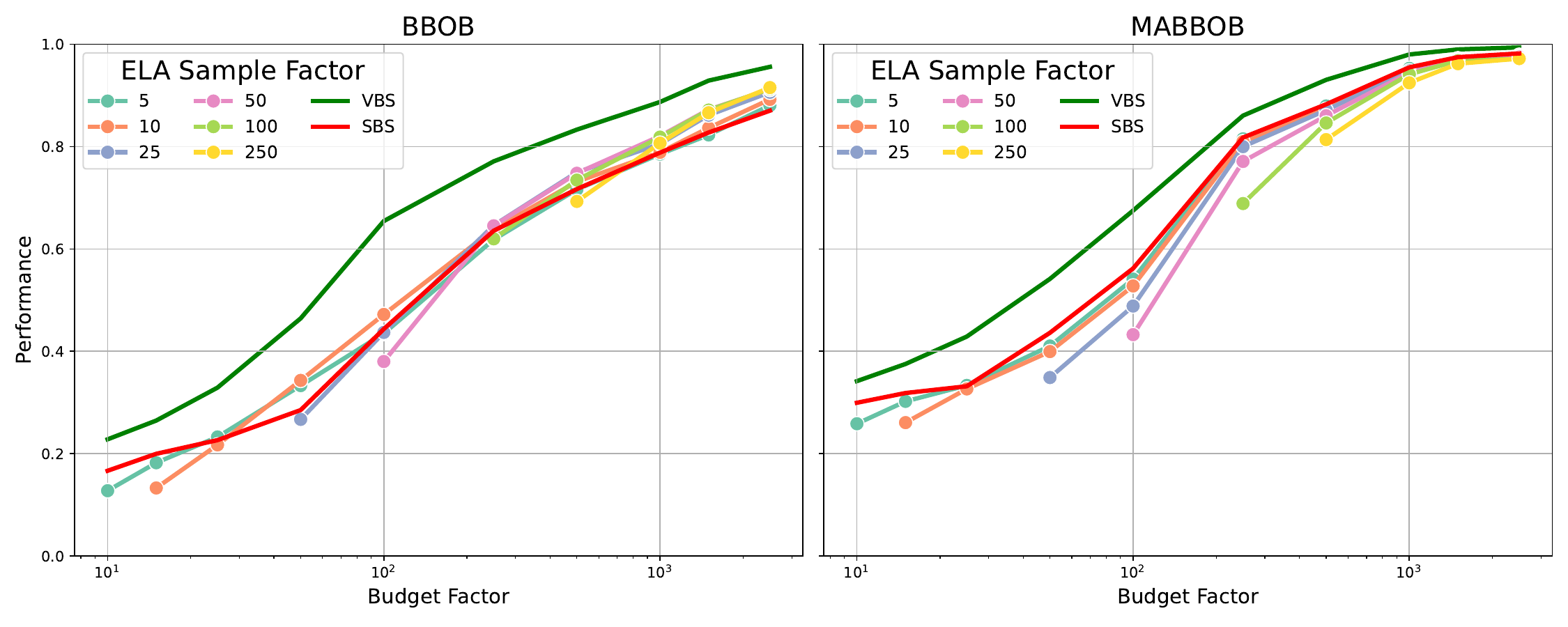}
    \caption{Performance of PIAS (mean over 5 folds) with different feature computation budgets compared to the VBS and SBS baselines for the 2-dimensional BBOB (left) and MA-BBOB (right) problems. Both plots use the full algorithm portfolios. }
    \label{fig:raw_gapclosed}
\end{figure}

The changes of PIAS effectiveness relative to the portfolio size are clear to see in Figure~\ref{fig:Other_gapclosed}. There are two potential factors that influence this. On the one hand, the smaller portfolios have been created in order to maximize algorithm complementarity (size of the SBS-VBS gap), which makes it easier for a decent selection process to improve over the SBS. On the other hand, smaller portfolios tend to be easier for selection methods to handle; the regression-based model we use might be less impacted by compounding errors in predicting too many output nodes.

When we look at the MA-BBOB results in particular, we observed that the SBS on the full algorithm portfolio is very strong. In Figure~\ref{fig:raw_gapclosed}, rather than the gap-closed measure, we show the actual performance of $\PIAS$ relative to the $\SBS$ (for a different scenario, with $d=2$), and notice the clear differences in the size of the SBS-VBS gap between the two scenarios. For MA-BBOB with the full algorithm portfolio (on the right), there are many problem instances where the SBS performs very well, and a limited number of instances where other generally poor-performing methods are much more suitable. Given the limited training set size, it seems to be challenging for our approach to accurately predict which instances fall in these categories, resulting in poor performance on the testing sets. This also provides additional support for the large influence the algorithm portfolio can have on the difficulty of an AS scenario, in addition to the influence of the considered problem set.

\begin{figure}[h!]
    \centering
    \includegraphics[width=0.45\linewidth]{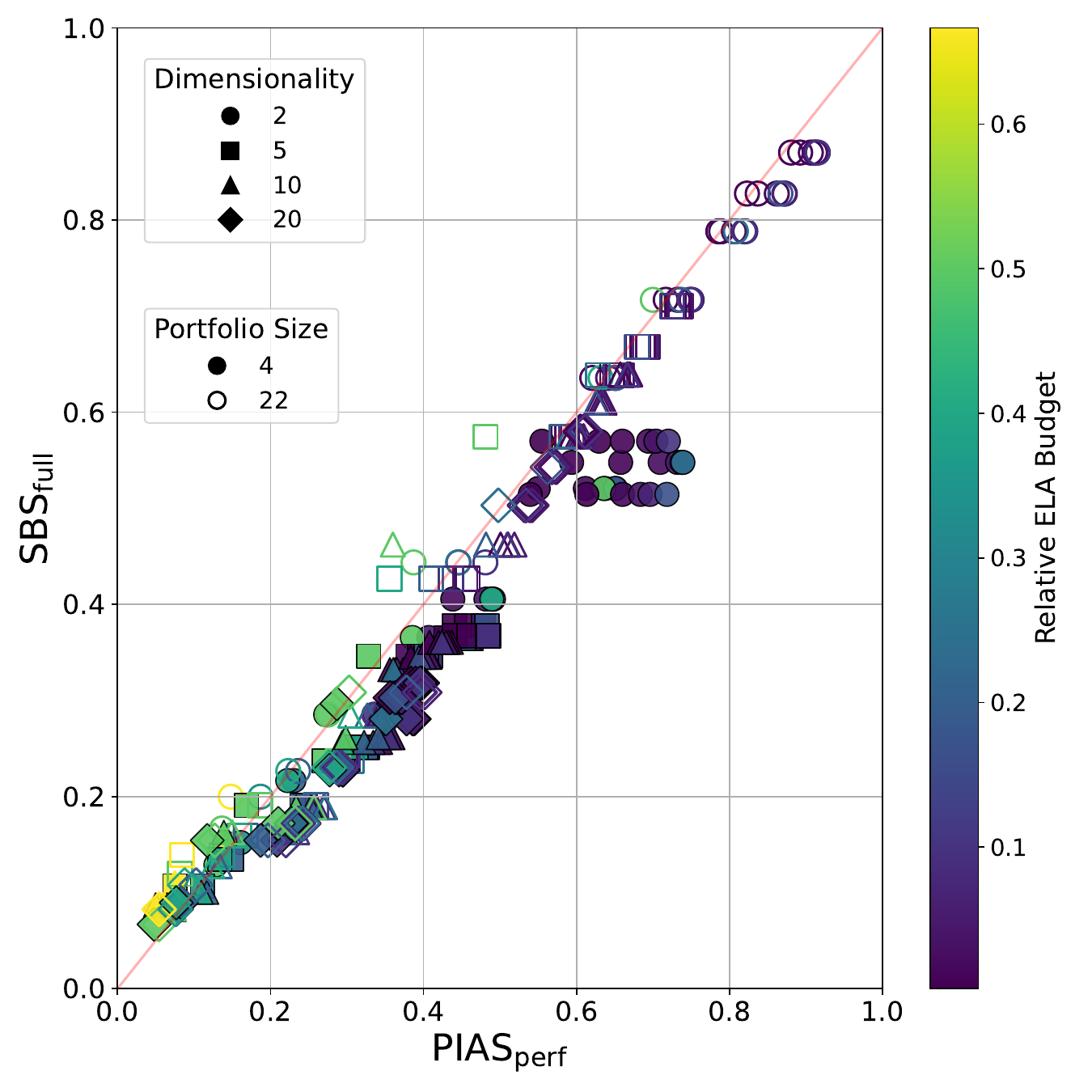}
    \includegraphics[width=0.45\linewidth]{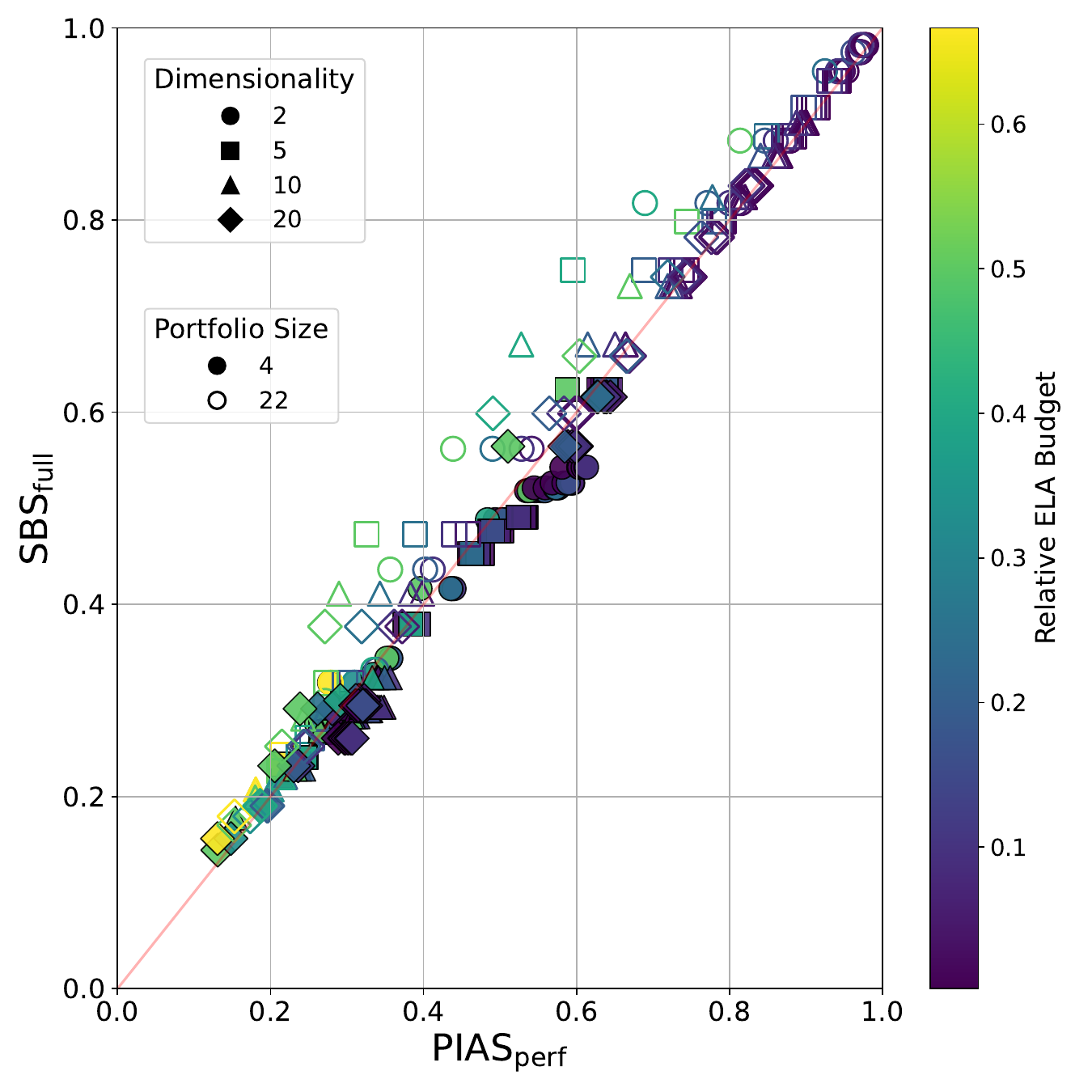}
    \caption{Comparison of $\SBS$ (y-axis) and $\PIAS$ (x-axis) results over all settings (mean over 5 folds), for BBOB (left) and MA-BBOB (right) respectively. Color indicates the fraction of budget used for feature calculation. Each dot corresponds to one AS scenario: budget, feature budget, dimensionality. Both portfolio sizes are included. Dots below the diagonal indicate cases where PIAS is beneficial.}
    \label{fig:PIAS_SBS_Scatter}
\end{figure}

To identify when spending a fraction of the budget on feature computation is worthwhile, we looked at the gap-closed by PIAS, which provides a relative measure of performance.
By considering only the comparison of the raw performance of the SBS and PIAS, we get a better view of the practical difference. Since we have many scenarios for each problem suite, we take a high-level view of this tradeoff across all scenarios. In Figure~\ref{fig:PIAS_SBS_Scatter}, we show the difference in performance between the SBS and PIAS for both the BBOB and MA-BBOB suites. This overview confirms the general view that the BBOB scenarios seem to benefit much more clearly from per-instance selection, while the tradeoff is much more mixed in the MA-BBOB setting. An explanation for this is that the BBOB problems create a scenario with well-separated clusters of instances belonging to each of the component problems (essentially the prototypical algorithm selection scenario), whereas the MA-BBOB scenarios consist of randomly generated instances without such obvious clusters (potentially a much harder scenario).
We also note that the results here in Figure~\ref{fig:PIAS_SBS_Scatter} reflect what we previously observed in Figure~\ref{fig:Other_gapclosed}: for MA-BBOB, PIAS almost always failed to be beneficial for the size-22 algorithm portfolio, whereas for the size-4 portfolio (with sufficient budget), it often achieved at least small gains. 

\subsection{Error Decomposition}

\begin{figure}[t]
    \centering
    \includegraphics[width=0.45\linewidth]{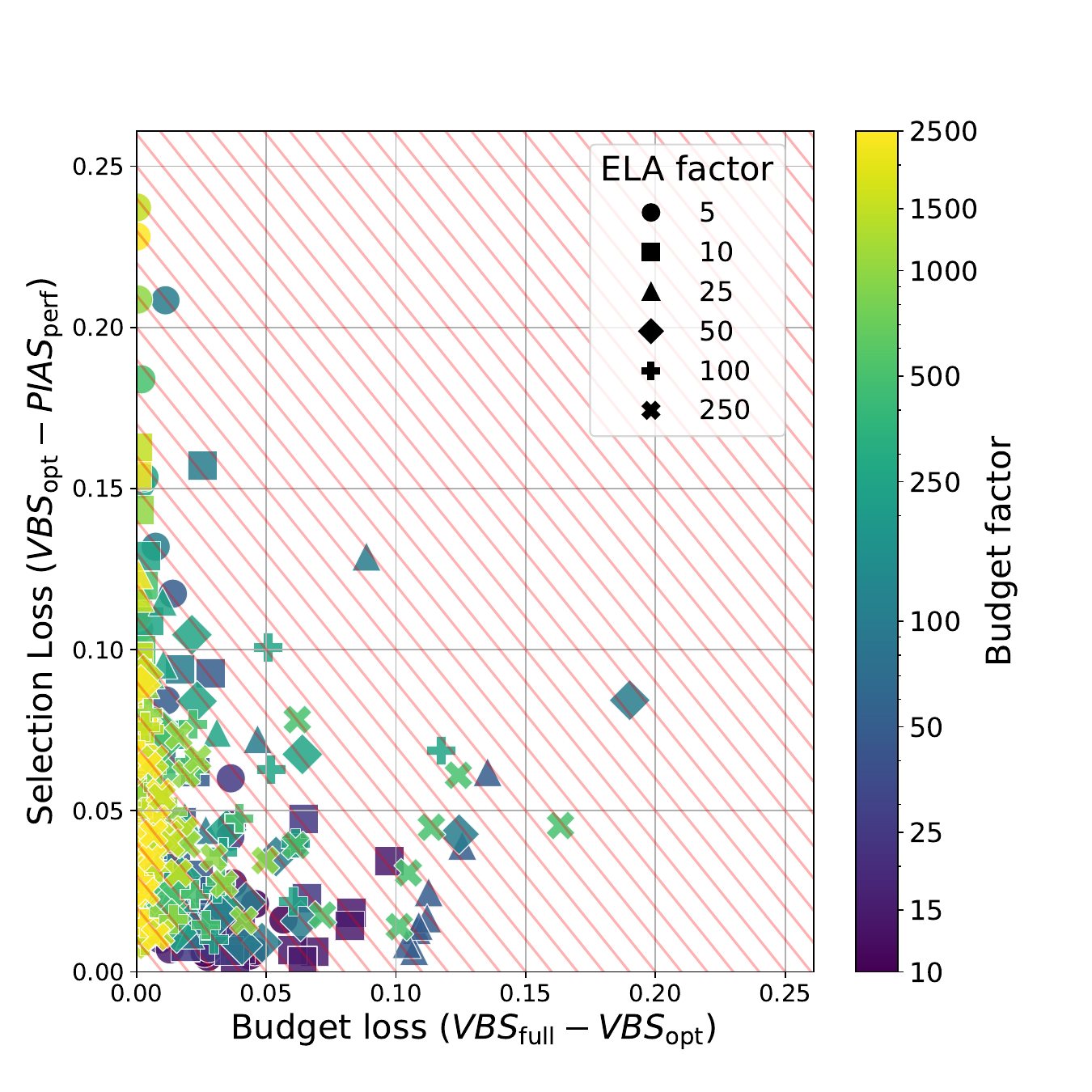}
    \includegraphics[width=0.45\linewidth]{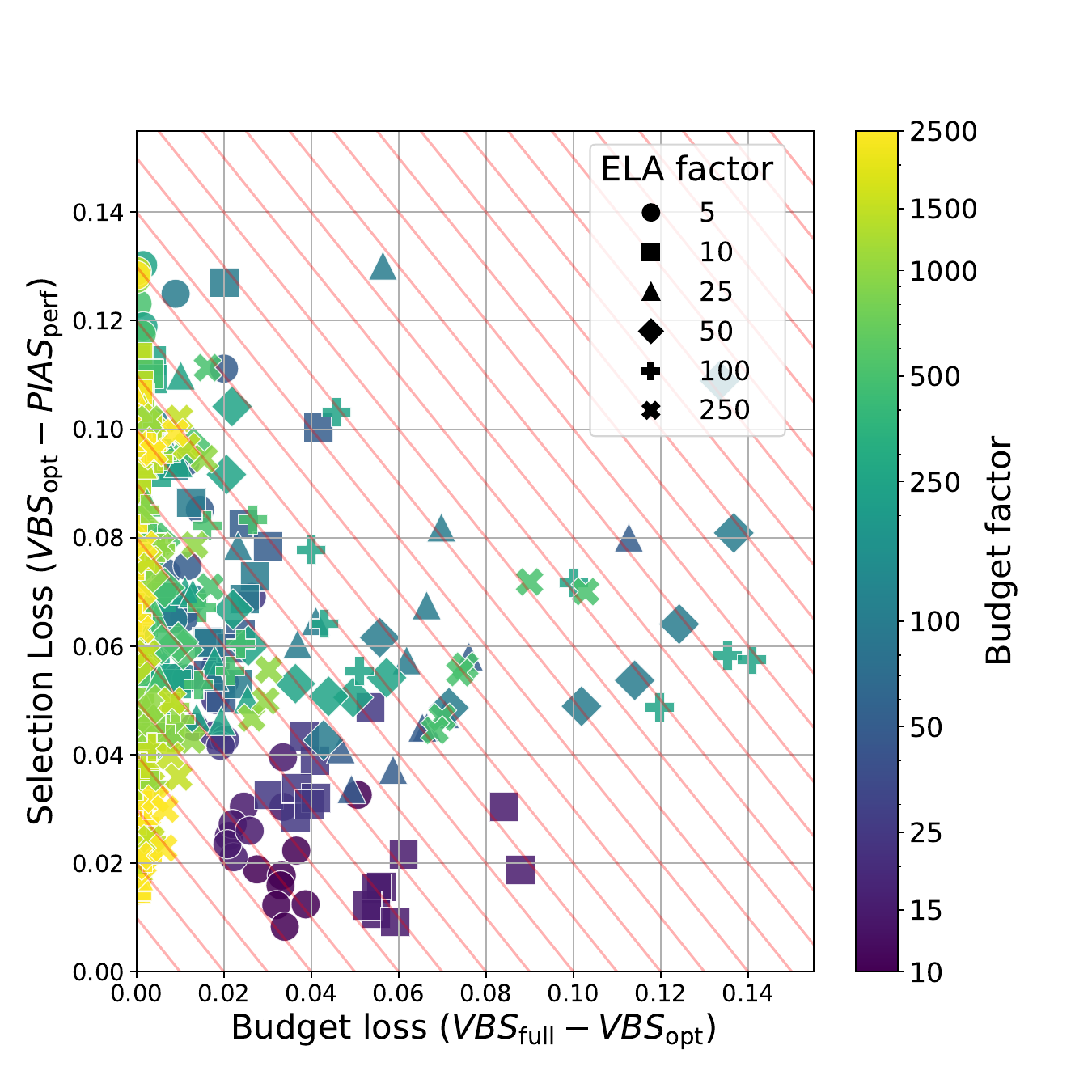}
    \caption{Decomposition of the $\PIAS$ loss (mean over 5 folds) into the loss that is occurring based on the selection procedure itself ($\VBSopt-\PIAS$, y-axis), and the loss resulting from the budget needed for feature calculation ($\VBS-\VBSopt$, x-axis). Both the full portfolio and the one with 4 algorithms are included. Each dot represents a single algorithm selection scenario. Left: BBOB. Right: MA-BBOB.
    }
    \label{fig:decomposition}
\end{figure}

To better understand the source of the loss resulting from using different fractions of the budget for feature computation, we lastly look at the two factors that make up the performance of PIAS. On the one side, there is the potential loss in performance the selected algorithm has when it can not use the full budget, because part of it is used for feature computation. The other factor is that the selector should more reliably choose better algorithms as the features more accurately represent the instances when larger feature computation budgets are used~\cite{renau2019expressiveness}. Given this distinction, we decompose the loss of the algorithm selector (how much it loses over the $\VBS$) into the selection-based loss (how much the selector, $\PIAS$, loses relative to the $\VBSopt$) and the Budget-based loss (how much the $\VBSopt$ loses relative to the $\VBS$).
In Figure~\ref{fig:decomposition}, we show this decomposition for the BBOB and MA-BBOB problem sets.

\begin{figure}[t]
    \centering
    \includegraphics[width=0.95\linewidth]{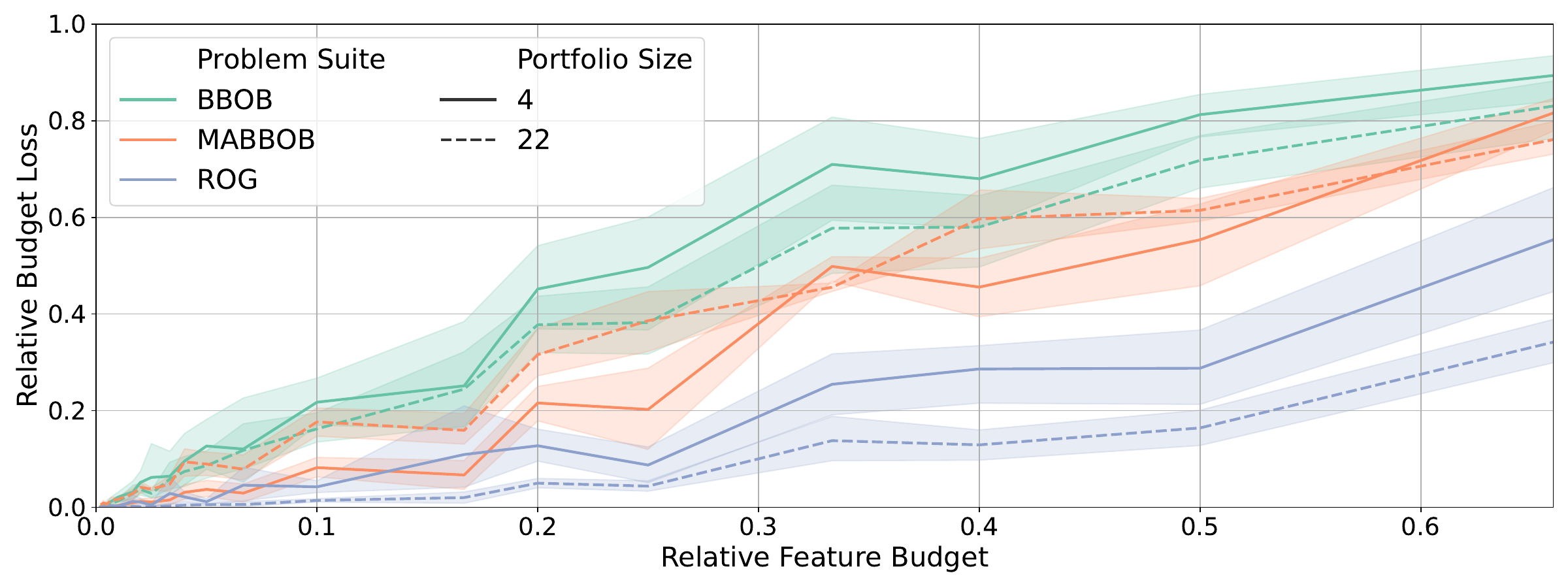}
    \caption{Average and $95\%$ confidence interval of relative budget loss ($\frac{\VBS-\VBSopt}{\VBS-\PIAS}$) across all algorithm selection scenarios and folds, as a function of the relative evaluation budget given to the feature computation. }
    \label{fig:relative_budget_loss}
\end{figure}

From Figure~\ref{fig:decomposition}, we see that generally, most of the loss comes from the selection procedure itself. There are also clear patterns present with respect to the optimization budget, which seems to be the main factor that determines budget loss. In order to get a better view of how this relation between budget and selection loss differs based on the budget setting, we can plot it as a function of the relative feature computation budget, as is done in Figure~\ref{fig:relative_budget_loss}. From this figure, we see a clear upward trend as a larger fraction of the budget is spent on feature computation, for all problem settings. The differences between problem suites are also clearly visible. When averaged over all algorithm selection scenarios for each problem suite, we see that the budget-based loss makes up $11, 22$ and \SI{28}{\percent} of the total loss for ROG, MA-BBOB, and BBOB, respectively.  This shows that even for settings where the selection works well, the change in optimization budget remains impactful and prevents the algorithm selection from reaching the performance of the $\VBS$. 

\section{Conclusions}

Per-instance algorithm selection (PIAS) has been successfully applied in domains such as SAT~\cite{XuEtAl2008}, MIP~\cite{XuEtAl2011}, and black-box optimization (BBO)~\cite{BischlEtAl2012}. In BBO, part of the total evaluation budget has to be spent on sampling to compute the instance-features that are needed to apply PIAS. In this work, we investigated the impact on PIAS performance of the fraction of the total budget spent on feature computation. To this end, we also accounted for the performance of the solutions sampled for feature computation, which may especially have an influence for smaller total budgets. To get a broad view, our experiments covered a wide range of in total \num{1440} algorithm selection scenarios, including three problem suites (BBOB~\cite{HansenEtAl2009}, MA-BBOB~\cite{VermettenEtAl2025}, RandOptGen~\cite{SeilerEtAl2025b}), two algorithm portfolio sizes (4, 22), four problem dimensionalities, ten total budgets, and six feature extraction budgets.

We have shown that the budget spent on feature computation can not just be ignored when reporting PIAS performance in the context of BBO. On average across problem suites, we observe that between 11 and \SI{28}{\percent} of the final loss of PIAS is caused by having to use a fraction of the evaluation budget for feature computation. While this fraction is highly dependent on the relation between feature computation budget and total budget, it is not negligible, and should thus be properly reported in any algorithm selection study where sampling is required for feature computation. 

A core focus of this paper has been to identify when it is worthwhile to use PIAS over sticking with the single best solver (SBS). In this regard, we observed quite some differences between the problem suites used. For scenarios with limited complementarity between optimizers, the SBS can be quite powerful, which leads to the selector performing worse given its lower total budget. However, in the standard BBOB scenario, where problems are complementary by design, we see that even low feature budgets lead to very good selector performance. We believe our results on a broad range of AS scenarios serve as additional evidence that, indeed, BBOB is too easy to meaningfully differentiate between the performance of AS techniques. Even so, we still see value in its use for AS to verify whether new ideas work at all, and, under the right conditions, to perform deeper analysis on specific components of the AS pipeline.

In general, we find that PIAS is quite beneficial in most of the scenarios we considered, with the main determining factors being the complementarity between algorithms in the portfolio and the fraction of the total budget spent on feature computation. Even with as much as a quarter of the budget spent on feature calculation, PIAS can outperform the SBS. However, without proper evaluation procedures, there is a risk of overstating the benefits of PIAS for BBO. As such, standard benchmarking practices for PIAS need to properly account for this particularity of algorithm selection in a black-box optimization context. 

\section{Future work}

Having looked at a large variety of algorithm selection scenarios, we have seen that the difficulty of the selection scenario is influenced by many different factors. This includes, for example, the algorithm portfolio and the instance set. It remains challenging to fully untangle the impact of each of these factors on the selection procedure. As such, continued development of tools to properly capture both the difficulty of algorithm selection scenarios and the performance of algorithm selectors themselves remains an important research direction. 

Our results show that in some cases, the performance achieved from sampling for feature calculation can be better than that of the selected algorithms. Such cases highlight the need for properly integrating the feature computation into the analysis of the algorithm selection results. This does not have to be limited to taking the maximum of the two, but could include warm-starting the selected algorithm, as suggested in, e.g.,~\cite{KerschkeTrautmann2019}. 

This aspect of warm-starting then naturally leads to a more dynamic algorithm selection scenario, where instead of relying on random sampling for feature computation, an initial optimization algorithm is run and its trajectory is used for feature computation~\cite{jankovic2021towards,RenauHart2024}. Initial studies into this \textit{dynamic algorithm selection} setting have shown promising results~\cite{KostovskaEtAl2022}, but further research is needed to validate its usability in a broader context. 

Finally, our results indicate a need for the creation of more testbeds for algorithm selection in the black-box optimization context. While the suites used in our paper are able to generate a variety of algorithm selection scenarios, selection-specific settings, such as algorithm complementarity in both problem and algorithm space, are hard to control. Having more robust ways to create algorithm selection scenarios with given properties would be greatly beneficial towards further study of the different components of PIAS in the BBO context.

\begin{credits}
\subsubsection{\ackname}
Koen van der Blom acknowledges funding by NWO WISE grant number 24.0742 ``Transfer Learning for Evolutionary Algorithms applied to Optimisation Problems''. 
Diederick Vermetten acknowledges funding by the European Union (ERC, ``dynaBBO'', grant no.~101125586). 
\subsubsection{\discintname}
The authors have no competing interests to declare that are
relevant to the content of this article.

\end{credits}

\bibliographystyle{splncs04}
\bibliography{bib}

\end{document}